\documentclass{article}

\usepackage[nonatbib, final]{neurips_2022}

\usepackage{microtype}
\usepackage{graphicx}
\usepackage{subfigure}
\usepackage{booktabs} %

\usepackage[pdfa]{hyperref}
\hypersetup{      
    pdftitle = {Deep Differentiable Logic Gate Networks},
    pdflang = {en-US},
    bookmarks = true,
}
\usepackage{url}

\usepackage[backend=biber, style=ieee]{biblatex}
\usepackage{wrapfig}

\usepackage{enumitem}
\setlist{leftmargin=2em,topsep=-2pt,itemsep=-.5ex,partopsep=1ex,parsep=1ex}

\usepackage[font=small]{caption}

\usepackage{tikz}

\newcommand{\etal}[0]{\textit{et al.}}

\newcommand{\citet}[1]{{\color{cyan}\textbf{CITET:} #1}}
\newcommand{\citep}[1]{{\color{cyan}\textbf{CITEP:} #1}}

\usepackage{amsmath}
\usepackage{xcolor}

\usepackage{amsmath,amsfonts,bm}

\def\eqref#1{equation~\ref{#1}}

\def\1{\bm{1}}

\def\vp{{\bm{p}}}

\def\vw{{\bm{w}}}

\DeclareMathAlphabet{\mathsfit}{\encodingdefault}{\sfdefault}{m}{sl}
\SetMathAlphabet{\mathsfit}{bold}{\encodingdefault}{\sfdefault}{bx}{n}

\def\sR{{\mathbb{R}}}

\DeclareMathOperator*{\argmax}{arg\,max}

\usepackage{amssymb}%
\usepackage{pifont}%

\usepackage{amsthm}
\newtheorem{theorem}{Theorem}

\newtheorem{definition}[theorem]{Definition}

\newtheorem{remark}[theorem]{Remark}

\title{Deep Differentiable Logic Gate Networks}
\author{%
  Felix Petersen\\
  Stanford University\\
  University of Konstanz\\
  \texttt{mail@felix-petersen.de}\\
  \And
  \\[-.5em]
  \textbf{Christian Borgelt}\\
  University of Salzburg\\
  \!\texttt{christian@borgelt.net}\!\\
  \AND
  Hilde Kuehne\\
  University of Frankfurt\\ %
  MIT-IBM Watson AI Lab\\
  \texttt{kuehne@uni-frankfurt.de}\\
  \And
  \\[-.5em]
  \textbf{Oliver Deussen}\\
  University of Konstanz\\
  \texttt{oliver.deussen@uni.kn}\\
}

\begin{document}

\maketitle

\begin{abstract}

Recently, research has increasingly focused on developing efficient neural network architectures. In this work, we explore logic gate networks for machine learning tasks by learning combinations of logic gates. These networks comprise logic gates such as ``AND'' and ``XOR'', which allow for very fast execution. The difficulty in learning logic gate networks is that they are conventionally non-differentiable and therefore do not allow training with gradient descent. Thus, to allow for effective training, we propose differentiable logic gate networks, an architecture that combines real-valued logics and a continuously parameterized relaxation of the network. The resulting discretized logic gate networks achieve fast inference speeds, e.g., beyond a million images of MNIST per second on a single CPU core.

\end{abstract}

\section{Introduction}

With the success of neural networks, there has also always been strong interest in research and industry in making the respective computations as fast and efficient as possible, especially at inference time. 
Various techniques have been proposed to solve this problem, including reduced computational precision \cite{choi2018pact, gupta2015deep}, binary \cite{qin2020binary} and sparse \cite{hoefler2021sparsity} neural nets.
In this work, we want to train a different kind of architecture, which is well known in the domain of computer architectures: logic (gate) networks.

The problem in training networks of discrete components like logic gates, is that they are non-differentiable and therefore, conventionally, cannot be optimized via standard methods such as gradient descent \cite{rumelhart1986learning}. 
One approach for this would be gradient-free optimization methods such as evolutionary training \cite{telikani2021evolutionary, nevergrad}, which works for small models, but becomes infeasible for larger ones.

In this work, we propose an approach for gradient-based training of logic gate networks (aka.~arithmetic / algebraic circuits~\cite{darwiche2002knowledge, arora2009computational}). 
Logic gate networks are based on binary logic gates, such as ``and'' and ``xor'' (see Table~\ref{tab:operators}).
For training logic gate networks, we continuously relax them to differentiable logic gate networks, which allows efficiently training them with gradient descent.
For this, we use real-valued logic and learn which logic gate to use at each neuron. 
Specifically, for each neuron, we learn a probability distribution over logic gates.
After training, the resulting network is discretized to a (hard) logic gate network by choosing the logic gate with the highest probability.
As the (hard) logic gate network comprises logic gates only, it can be executed very fast.
Additionally, as the logic gates are binary, every neuron / logic gate has only 2 inputs, and the networks are extremely sparse. %

Logic gate networks are not binary neural networks: binary neural networks are a form of low precision (wrt.~weights and/or activations) neural networks, as they reduce weights and/or activations to binary precision.
Binary neural networks are usually dense and typically rely on weights trained in the continuous domain and are discretized afterwards. 
In contrast to binary neural networks, logic gate networks do not have weights, are intrinsically sparse as they have only 2 inputs to each neuron, and are not a form of low precision (wrt.~weights and/or activations) neural networks.

They also differ from current sparse neural network approaches, as our goal is to learn which logic gate operators are present at each neuron, while the (weightless) connections between neurons are (pseudo-)randomly initialized and remain fixed.
The network is, thus, parameterized by the choice of the logic gate operator / binary function for each neuron. 
As there is a total of $16$ functions of signature $f: \{0,1\}\times\{0,1\}\to\{0,1\}$, the information which operation a neuron executes can be encoded in just $4$ bits.
The objective is to learn which of those $16$ operations is optimal for each neuron.
Specifically, for each neuron, we learn a probability distribution over possible logic gates, which we parameterize via softmax.
We find that this approach allows learning logic gate networks very effectively via gradient descent.

\begin{figure*}[t]
    \centering
    \includegraphics[width=\linewidth]{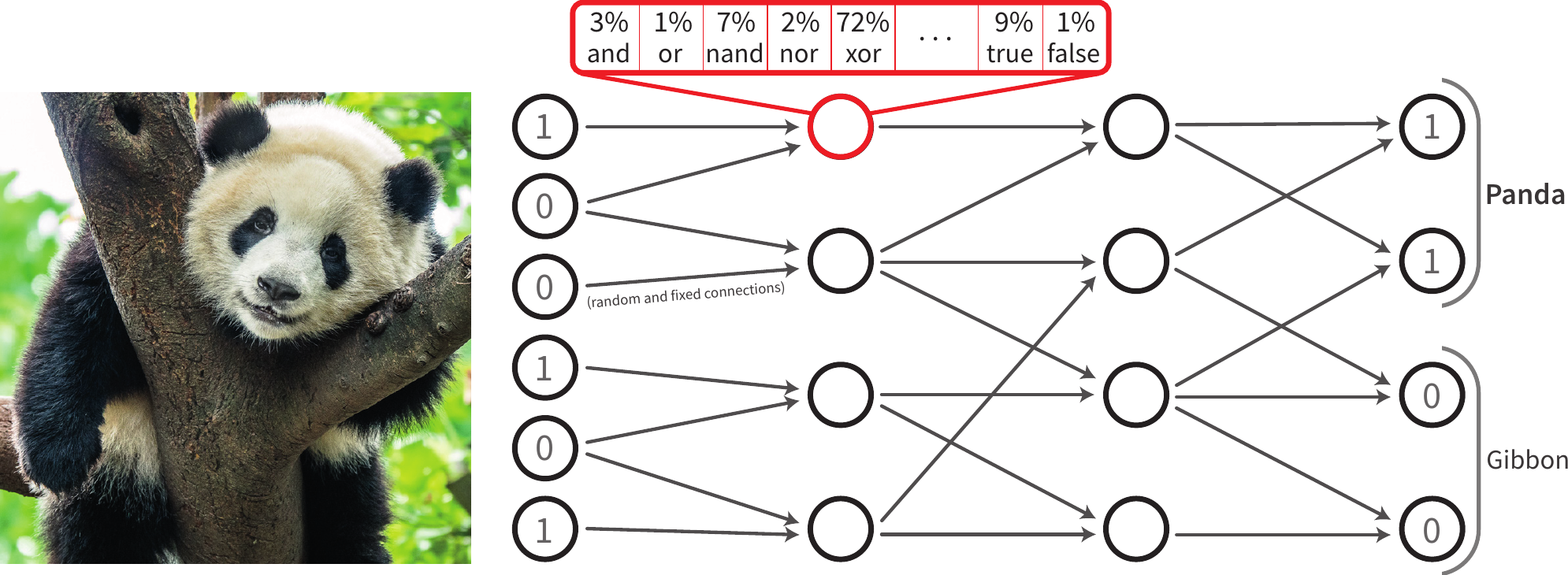}
    \vspace{-1em}
    \caption{
    Overview of the proposed differentiable logic gate networks: 
    the pixels of the image are converted into Boolean valued inputs, which are processed by a layer of neurons such that each neuron receives two inputs.
    The connectivity of neurons remains fixed after an initial pseudo-random initialization.
    Each neuron is continuously parameterized by a distribution over logical operators. 
    During training, this distribution is learned for each neuron, and, during inference, the most likely operator is used for each neuron.
    There are multiple outputs per class, which are aggregated by bit-counting, which yields the class scores. 
    The number of neurons in the visualization is greatly reduced for visual simplicity.
    }
    \vspace{-.5em}
    \label{fig:overview-small-net}
\end{figure*}

Logic gate networks allow for very fast classification, with speeds beyond a million images per second on a single CPU core (for MNIST at $>97.5\%$ accuracy).
The computational cost of a layer with $n$ neurons is $\Theta(n)$ with very small constants (as only logic gates of Booleans are required), while, in comparison, a fully connected layer (with $m$ input neurons) requires $\Theta(n\cdot m)$ computations with significantly larger constants (as it requires floating-point arithmetic).
While the training can be more expensive than for regular neural networks (however, just by a constant and asymptotically less expensive), to our knowledge, the proposed method is the fastest available architecture at inference time.
Overall, our method accelerates inference speed (in comparison to fully connected ReLU neural networks) by around two orders of magnitude. 
In the experiments, we scale the training of logic gate networks up to 5 million parameters, which can be considered relatively small in comparison to other architectures.
In comparison to the fastest neural networks at $98.4\%$ on MNIST, our method is more than $12\times$ faster than the best binary neural networks and $2-3$ orders of magnitude faster than the theoretical speed of sparse neural networks.

\section{Related Work}

\label{sec:difflogic:rel-work}

In this section, we discuss related work on learning logic gate networks, methods with methodological or conceptual similarity, as well as other machine learning methods that are fast and that we compare ourselves to with respect to inference cost and speed.

\paragraph{Differentiable Logics and Triangular Norms}

Differentiable logics (aka.~real-valued logics, or infinite-valued logic) are well-known in the fields of fuzzy logics~\cite{klir1997fuzzy} and probabilistic metric spaces~\cite{menger1942statistical, klement2013triangular}.
In Supplementary Material~\ref{sm:tcn} we give examples for T-norms and T-conorms.
An additional reference for differentiable real-valued logics is Van~\etal~\cite{van2020analyzing}.

\paragraph{Learning Logic Gate Networks}

Chatterjee~\cite{chatterjee2018learning} explored ``memorization'', a method for memorizing binary classification data sets with a network of binary lookup tables.
The motivation for this is to explore principles of learning and memorization, as well as their trade-off and generalization capabilities.
He constructs the networks of lookup tables by counting conditional frequencies of data points.
We mention this work here because binary logic gates may be seen as the special case of $2$-input lookup tables.
That is, his method has some similarities to our resulting networks.
However, as he memorizes the data set, while this leads to some generalization, this generalization is limited.
In his experiments, he considers the binary classification task of distinguishing the combined classes `0'--`4' from the combined classes `5'--`9' of MNIST and achieves a test accuracy of $90\%$.
Brudermueller~\etal~\cite{brudermueller2020making} propose a method where they train a neural network on a classification task and then translate it, first into random forests, and then into networks of AND-Inverter logic gates, i.e., networks based only on ``and'' and ``not'' logical gates.
They evaluate their approach on the ``gastrointestinal bleeding'' and ``veterans aging cohort study'' data sets and argue for the verifiability and interpretability of small logical networks in patient care and clinical decision-making.

\paragraph{Continuous Relaxations}

A popular approach for making discrete structures differentiable is continuous relaxation~\cite{cuturi2019differentiable, petersen2022learning}. 
In this work, we also use continuous relaxations; however, instead of relaxing a fixed discrete structure (e.g., an algorithm)~\cite{cuturi2019differentiable, petersen2022learning, petersen2022gendr, petersen2022monotonic, berthet2020learning, petersen2021diffsort, cuturi2017soft, petersen2021learning}, we continuously relax a discrete structure (a logic gate network) to optimize it.

\paragraph{Relaxed Connectivity in Networks}

In this work, we relax \textit{which logic operator} is applied at each node, while the connections are predefined.
Zimmer~\etal~\cite{zimmer2021differentiable} propose differentiable logic machines for inductive logic programming.
For this, they propose logic modules, which contain one level of logic and for which they predefine that the first half of operators are fuzzy ``and''s and the second half are fuzzy ``or''s.
They relax \textit{which nodes} are the inputs to the ``and''s and ``or''s of their logic modules.
Similarly, Chen~\cite{chen2020learning} proposes Gumbel-Max Equation Learner Networks, where he predefines a set of arithmetic operations in each layer and learns via Gumbel-Softmax~\cite{maddison2017concrete, jang2017categorical}, \textit{which outputs} of the previous layer should be used as inputs of a respective arithmetic operation.
He uses this to learn symbolic expressions from data.
While these works relax which nodes are connected to which nodes, this is fixed in our work, and we relax which operator is at which node.

\paragraph{Evolutionary Learning of Networks}

Mocanu~\etal~\cite{mocanu2018scalable} propose training neural networks with sparse evolutionary training inspired by network science.
Their method evolves an initial sparse topology of two consecutive layers of neurons into a scale-free topology.
On MNIST, they achieve (with $89\,797$ parameters) an accuracy of $98.74\%$.
Gaier~\etal~\cite{gaier2019weight} propose learning networks of operators such as ReLU, sin, inverse, absolute, step, and tanh using evolutionary strategies.
Specifically, they use the population-based neuroevolution algorithm NEAT.
They achieve learning those floating-point function-based networks and achieve an accuracy of $94.2\%$  on MNIST with a total of $1\,849$ connections.

\paragraph{Learning of Decision Trees}

Zantedeschi~\etal~\cite{zantedeschi2021learning} propose to learn decision trees by quadratically relaxing the decision trees from mixed-integer programs that learn the discrete parameters of the tree (input traversal and node pruning).
This allows them to differentiate in order to simultaneously learn the continuous parameters of splitting decisions.
Logic gate-based trees are conceptually vastly different from decision trees:
decision trees rely on splitting decisions instead of logical operations, and the tree structure of decision trees and logic gate-based trees are in the opposite directions~\cite{clark1989cn2}.
Logic gate-based trees begin with a number of inputs (leafs) and apply logic gates to aggregate them to a binary value (root).
Decision trees begin at the root and apply splitting decisions (for which they consider an external input) to decide between children, such that they end up at a leaf node corresponding to a value. 

\paragraph{Binary Neural Networks}
Binary neural networks (BNNs)~\cite{qin2020binary} are conceptually very different from logic gate networks.
For binary neural networks, ``binary'' refers to representing activations and weights of a neural network with binary states (e.g., $\{-1, +1\}$).
This allows approximating the expensive matrix multiplication by faster XNOR and bitcount (popcount) operations.
The logical operations involved in BNNs are not learned but instead predefined to approximate floating-point operations, and, as such, a regular weight-based neural network.
This is not the case for logic gate networks, where we learn the logic operations, we do not approximate weight-based neural networks, and do not have weights.
While BNNs are defined via their weights and not via their logic operations, logic gate networks do not have weights and are purely defined via their logic operations.
We include BNNs as baselines in our experiments because they achieve the best inference speed.

\paragraph{Sparse Neural Networks}

Sparse neural networks~\cite{hoefler2021sparsity} are neural networks where only a selected subset of connections is present, i.e., instead of fully-connected layers, the layers are \textit{sparse}.
In the literature of sparse neural networks, usually, the task is to distill a sparse neural network from a dense neural network and the choice of connections is important.
However, there has also been work suggesting a high effectiveness of using randomized and fixed sparse connections~\cite{liu2022unreasonable}.
For logic gate networks, which are sparse by definition, we follow these findings and use randomly initialized and fixed connections. %

\section{Logic Gate Networks}

Logic gate networks are networks similar to neural networks where each neuron is represented by a binary logic gate like `and', `nand', and `nor' and accordingly has only two inputs (instead of all neurons in the previous layer as it is the case in fully-connected neural networks).
Given a binary vector as input, pairs of Boolean values are selected, binary logic gates are applied to them, and their output is then used as input for layers further downstream.
Logic gate networks do not use weights. 
Instead, they are parameterized via the choice of logic gate at each neuron.
In contrast to fully connected neural networks, binary logic gate networks are sparse because each neuron has only $2$ instead of $n$ inputs, where $n$ is the number of neurons per layer.
In logic gate networks, we do not need activation functions as they are intrinsically non-linear.

While it is possible to make a prediction simply with a single binary output or $k$ binary outputs for $k$ classes, this is not ideal. 
This is because in the crisp case, we only get $0$s or $1$s and no graded prediction, which would be necessary for a ``greatest activation'' classification scheme.
By using multiple neurons per class and aggregating them by summation, even the crisp case allows for grading, with as many levels as there are neurons per class.
Each of these neurons captures a different piece of evidence for a class, and this allows for more finely graded predictions.
Figure~\ref{fig:overview-small-net} illustrates a small logic gate network.
In the illustration, each node corresponds to a single logic operator. 
Note that the distribution over operators (red) is part of the differentiable relaxation discussed in the next section.

As logic gate networks build on bit-wise logic operations only, their execution is very efficient.

\section{Differentiable Logic Gate Networks}

Training binary logic gate networks is hard because they are not differentiable, and thus no gradient descent-based training is conventionally possible.
Thus, we propose relaxing logic gate networks to differentiable logic gate networks to allow for gradient-based training.

\paragraph{Differentiable Logics} 
\begin{wraptable}[17]{r}{.51\linewidth}
    \vspace{-1.75em}
    \centering
    \caption{
        List of all real-valued binary logic ops.%
        \label{tab:operators}
    }
    \scalebox{.94}{\footnotesize
    \addtolength{\tabcolsep}{-3pt}
    \begin{tabular}{rclccccccccc}
        \toprule
        ID    & Operator                    & real-valued   & 00 & 01 & 10 & 11 \\
        \midrule
           0  & False                       & $0$                       & 0     & 0     & 0     & 0     \\
           1  & $A\land B$                  & $A\cdot B$                & 0     & 0     & 0     & 1     \\
           2  & $\neg(A \Rightarrow B)$     & $A-AB$                    & 0     & 0     & 1     & 0     \\
           3  & $A$                         & $A$                       & 0     & 0     & 1     & 1     \\
           4  & $\neg(A \Leftarrow B)$      & $B-AB$                    & 0     & 1     & 0     & 0     \\
           5  & $B$                         & $B$                       & 0     & 1     & 0     & 1     \\
           6  & $A \oplus B$                & $A + B - 2AB$             & 0     & 1     & 1     & 0     \\
           7  & $A \lor B$                  & $A + B - AB$              & 0     & 1     & 1     & 1     \\
           8  & $\neg(A \lor B)$            & $1 - (A + B - AB)$        & 1     & 0     & 0     & 0     \\
           9  & $\neg(A \oplus B)$          & $1 - (A + B - 2AB)$       & 1     & 0     & 0     & 1     \\
           10 & $\neg B$                    & $1 - B$                   & 1     & 0     & 1     & 0     \\
           11 & $A \Leftarrow B$            & $1-B+AB$                  & 1     & 0     & 1     & 1     \\
           12 & $\neg A$                    & $1-A$                  & 1     & 1     & 0     & 0     \\
           13 & $A \Rightarrow B$           & $1-A+AB$                  & 1     & 1     & 0     & 1     \\
           14 & $\neg(A \land B)$           & $1 - AB$                  & 1     & 1     & 1     & 0     \\
           15 & True                        & $1$                       & 1     & 1     & 1     & 1     \\
        \bottomrule
    \end{tabular}
    }
\end{wraptable}

To make binary logic networks differentiable, we leverage the following relaxation. 
First, instead of hard binary activations / values $a \in \{0,1\}$, we relax all values to probabilistic activations $a \in [0,1]$.
Second, we replace the logic gates by computing the expected value of the activation given probabilities of independent inputs $a_1$ and $a_2$.
For example, the probability that two independent events with probabilities $a_1$ and $a_2$ both occur is $a_1\cdot a_2$. 
These operators correspond to the probabilistic T-norm and T-conorm;
we report the full set of relaxations corresponding to the probabilistic interpretation in Table~\ref{tab:operators}.
(In addition, we report alternative relaxations corresponding to alternative interpretations in Tables~\ref{tab.t-norms} and~\ref{tab.t-conorms} in Supplementary Material~\ref{sm:tcn}.)

Accordingly, we define the activation of a neuron with the $i$th operator as 
\begin{equation}
    a' = f_i(a_1, a_2)\,,
\end{equation}
where $f_i$ is the $i$th real-valued operator corresponding to Table~\ref{tab:operators} and $a_1, a_2$ are the inputs to the neuron.
There are also alternative real-valued logics like the Hamacher T-(co)norm, the relativistic Einstein sum, and the Łukasiewicz T-(co)norm.
While, in this work, we use the probabilistic interpretation, we review an array of possible T-norms and T-conorms that could also be used in SM~\ref{sm:tcn}.

\paragraph{Differentiable Choice of Operator} While real-valued logics allow differentiation, they do not allow training as the operators are not continuously parameterized and thus (under hard binary inputs) the activations in the network will always be $a \in \{0, 1\}$.
Thus, we propose to represent the choice of \textit{which} logic gate is present at each neuron by a categorical probability distribution.
For this, we parameterize each neuron with $16$ floats (i.e., $\vw\in\sR^{16}$), which, by softmax, map to the probability simplex (i.e., a categorical probability distribution such that all entries sum up to $1$ and it has only non-negative values).
That is, $\vp_i = e^{\vw_i} / (\sum_j e^{\vw_j})$, and thus $\vp$ lies in the probability simplex $\vp\in\Delta^{15}$.
During training, we evaluate for each neuron all $16$ relaxed binary logic gates and use the categorical probability distribution to compute their weighted average.
Thus, we define the activation $a'$ of a differentiable logic gate neuron as 
\begin{equation}
    a' = \sum_{i=0}^{15} \vp_i \cdot {f}_i(a_1, a_2) = \sum_{i=0}^{15} \frac{e^{\vw_i}}{\sum_j e^{\vw_j}} \cdot {f}_i(a_1, a_2)\,.
\end{equation}

\paragraph{Aggregation of Output Neurons}
Now, we may have $n$ output neurons $a_1, a_2, ..., a_n\in[0,1]$, but we may want the logic gate network to only predict $k<n$ values of a larger range than $[0,1]$. 
Further, we may want to be able to produce graded outputs.
Thus, we can aggregate the outputs as 
\begin{equation}
    \hat y_i = {\sum_{j=i\cdot n/k + 1}^{(i+1)\cdot n/k}}  a_j\, /\, \tau + \beta\,
\end{equation}
where $\tau$ is a normalization temperature and $\beta$ is an optional offset.

\subsection{Training Considerations}

\paragraph{Training}
For learning, we randomly initialize the connections and the parameterization of each neuron.
For the initial parameterization of each neuron, we draw elements of $\vw$ independently from a standard normal distribution.
In all reported experiments, we use the same number of neurons in each layer (except for the input) and between $4$ and $8$ layers, which we call straight network.
We train all models with the Adam optimizer \cite{kingma2015adam} at a constant learning rate of $0.01$.

\paragraph{Discretization}
After training, during inference, we discretize the probability distributions by only taking their mode (i.e., their most likely value), and thus the network can be computed with Boolean values, which makes inference very fast.
In practice, we observe that most neurons converge to one logic gate operation; therefore, the discretization step introduces only a small error, e.g., for MNIST, the gap is smaller than $0.1\%$. 
We note that all reported results are accuracies after discretization.

\paragraph{Classification}
In the application of a classification learning setting with $k$ classes (e.g., $10$) and $n$ output neurons (e.g., $1\,000$), we group the output into $k$ groups of size $n/k$ (e.g., $100$).
Then, we count the number of $1$s which corresponds to the classification score such that the predicted class can be retrieved via the $\argmax$ of the class scores.
During differentiable training, we sum up the probabilities of the outputs in each group instead of counting the $1$s, and we can train the model using a softmax cross-entropy classification loss.
For a reference on choosing the hyperparameter $\tau$, see Supplementary Material~\ref{sm:model-architectures-hyperparameters}; the offset $\beta$ is not relevant for the classification setting, as $\argmax$ is shift-invariant.
A heuristic for choosing $\tau$ is that when increasing $n$, we have to reduce $\tau$. 
Empirically, when increasing $n$ by a factor of $10$, $\tau$ should be decreased by a factor of about $2$ to $\sqrt{10}$.

\paragraph{Regression}
For regression learning, let us assume that we need to predict a $k$-dimensional output vector.
Here, $\tau$ and $\beta$ play the role of an affine transformation to transform the range of possible predictions from $0$ to $n/k$ to an application specific and more suitable range.
Here, the optional bias $\beta$ is important, e.g., if we want to predict values outside the range of $[0, n/k/\tau]$.
In some cases, it is desirable to cover the entire range of real numbers, which may be achieved using a logit transform $\operatorname{logit}(x)=\sigma^{-1}(x) = \log \frac{x}{1-x}$ in combination with $\tau=n/k, \beta=0$.
During differentiable training, we sum up the probabilities of the outputs in each group instead of counting the $1$s, and we can train the model, e.g., using an MSE loss.

\subsection{Remarks}

\paragraph{Boolean Vectorization via Larger Data Types}
One important computational detail for inference time is that we do not use Boolean data types but instead use larger data types such as, e.g., int64 for a batch size of $64$, and thus perform bit-wise logics on larger batches which significantly improves speed on current hardware. 
For int64, we batch $64$ data points such that the $i$th Boolean value of the $j$th data point is the $j$th bit in the $i$th int64 integer.
Thus, it is possible to compute on average around $250$ binary logic gates on each core in each CPU clock cycle (i.e., per Hz) on a typical desktop / notebook computer.
This is the case because modern CPUs execute many instructions per clock cycle even on a single core, and additionally (for Booleans) allow single-instruction multiple-data (SIMD) by batching bits from multiple data points into one integer (e.g., int64).
Using advanced vector extensions (AVX), even larger speedups would be possible.
On GPU, this computational speedup is also available in addition to typical GPU parallelization.

\paragraph{Aggregation of Output Neurons via Binary Adders}
In addition, during inference, we aggregate the output neurons directly using logic gate nets that make up respective adders, as writing all outputs to memory would constitute a bottleneck and aggregating them using logic gate networks is fast.
Specifically, we construct adders that can add exactly one bit to a binary number from logic gates.

\paragraph{Memory Considerations}
Since we pseudo-randomly initialize the connections in binary logic gate networks, i.e., which are the two inputs for each neuron, we do not need to store the connections as they can be reproduced from a single seed.
Thus, it suffices to store the $4$-bit information which of the $16$ logic gate operators is used for each neuron.
Thus, the memory footprint of logic gate networks is drastically reduced in comparison to neural networks, binary neural networks, and sparse neural networks.

\paragraph{Pruning the Model}
An additional speedup for the inference of logic gate networks is available by pruning neurons that are not used, or by simplifying logical expressions.
However, this requires storing the connections, posing a (minor) trade-off between memory and speed.

\paragraph{Subset of Operators}
We investigated reducing the set of operators; however, we found that, in all settings, the more expressive full set of $16$ operators performed better.
Nevertheless, a smaller set of operators could be a good trade-off for reducing the model size.

\paragraph{Half Precision}
We also investigated training with half precision (float16).
In our experiments, half precision (in comparison to full precision) did not degrade training performance; nevertheless, all reported results were trained with full precision (float32).

\paragraph{Optimizer} 
For training differentiable logic gate networks, we use the Adam optimizer~\cite{kingma2015adam} because it includes a normalization of the gradients with respect to their magnitude over past steps.
We found that this greatly improves training compared to other optimizers like SGD or SGD with momentum, which can become ineffective for training deeper logic gate networks.

\section{Current Limitations and Opportunities}

\paragraph{Expensive Training}
A limitation of differentiable logic gate networks is their relatively higher training cost compared to (performance-wise) comparable conventional neural networks.
The higher training cost is because multiple differentiable operators need to be evaluated for each neuron, and in their real-valued differentiable form, most of these operators require floating-point value multiplications.
However, the practical computational cost can be reduced through improved implementations.
We note that, asymptotically, differentiable logic gate networks are cheaper to train compared to conventional neural networks due to their sparsity.

\paragraph{Convolutions and Other Architectures}
Convolutional logic gate networks and other architectural components such as residual connections are interesting and important directions for future research.

\paragraph{Edge Computing and Embedded Machine Learning}
We would like to emphasize that the current limitations to rather small architectures (compared to large deep learning architectures) does not need to be a limitation: 
For example, in edge computing and embedded machine learning~\cite{murshed2021machine, ajani2021overview, seng2021embedded, branco2019machine}, models are already limited to tiny architectures because they run, e.g., on mobile CPUs, microcontrollers, or IoT devices.
In these cases, training cost is not a concern because it is done before deployment.

We also note that there are many other applications in industry where the training cost is negligible in comparison to the inference cost.

\section{Experiments\protect\footnote{The source code will be publicly available at \href{https://github.com/Felix-Petersen/difflogic}{\color{blue!50!black}github.com/Felix-Petersen/difflogic}.}}

To empirically validate our method, we perform an array of experiments.
We start with the three MONK data sets and continue to the Adult Census and Breast Cancer data sets.
For each experiment, we compare our method to other methods in terms of model memory footprint, evaluation speed, and accuracy.
To demonstrate that our method also performs well on image recognition, we benchmark it on the MNIST as well as the CIFAR-10 data sets.
We benchmark speeds and computational complexity of our method in comparison to baselines, which we discuss in detail in Section~\ref{sec:difflogic:baselines}.

\begin{wraptable}[14]{r}{0.7\linewidth}
    \centering
    \vspace{-3.5em}
    \caption{
        Results on the MONK data sets. The inference times are per data point for 1 CPU thread. Averaged over $10$ runs. 
        For Diff Logic Nets, $\#$ Parameters and Space vary between the MONK data sets as we use different architectures.
    }
    \scalebox{.94}{\footnotesize
    \addtolength{\tabcolsep}{-3pt}
    \begin{tabular}{lccc}
    \toprule
        Method & MONK-1 & MONK-2 & MONK-3 \\
    \midrule
        Decision Tree Learner (ID3) \cite{quinlan1986induction}     & $98.6\%$ & $67.9\%$ & $94.4\%$ \\
        Decision Tree Learner (C4.5) \cite{salzberg1994c4}     & $100\%$ & $70.4\%$ & $100\%$ \\
        Rule Learner (CN2) \cite{clark1989cn2}             & $100\%$ & $69.0\%$ & $89.1\%$ \\
        Logistic Regression             & $71.1\%$ & $61.4\%$ & $97.0\%$ \\
        Neural Network                  & $100\%$ & $100\%$ & $93.5\%$ \\
        Diff Logic Net (\textit{ours})   & $100\%$ & $90.9\%$ & $97.7\%$ \\
    \toprule
         & \# Parameters & Inf.~Time & Space \\
    \midrule
        Decision Tree Learner   & $\approx30$ & $49\textrm{ns}$ & $\approx60\textrm{B}$ \\
        Logistic Regression     & $20$  &   $68\textrm{ns}$     & $80\textrm{B}$ \\
        Neural Network          & $162$ &   $152\textrm{ns}$     & $648\textrm{B}$ \\
        Diff Logic Net (\textit{ours}) & $144\,|\,72\,|\,72$ & $18\textrm{ns}$ & $72\textrm{B}\,|\,36\textrm{B}\,|\,36\textrm{B}$ \\
    \bottomrule
    \end{tabular}
    }
    \label{tab:monk}
\end{wraptable}

\subsection{MONK's Problems}

The MONK's problems \cite{thrun1991monk} are 3 classic machine learning tasks that have been used to benchmark learning algorithms. 
They consist of 3 binary classification tasks on a data set with 6 attributes with $2-4$ possible values each.
Correspondingly, the data points can be encoded as binary vectors of size 17.
In Table~\ref{tab:monk}, we show the performance of our method, a regular neural network, and a few of the original learning methods that have been benchmarked.
We give the prediction speed for a single CPU thread, the number of parameters, and storage requirements.

\begin{wraptable}[12]{r}{0.56\linewidth}
    \centering
    \vspace{-.5em}
    \caption{Results for the Adult and Breast Cancer data sets averaged over 10 runs.}
    \scalebox{.94}{\footnotesize
    \addtolength{\tabcolsep}{-3pt}
    \begin{tabular}{lcccc}
    \toprule
        \textbf{Adult} & Acc. & \# Param. & Infer.~Time & Space \\
    \midrule
        Decision Tree Learner           & $79.5\%$ & \kern-1.5ex$\approx50$ & $86$\textrm{ns} & \kern-2.3ex$\approx130$B \\
        Logistic Regression             & $84.8\%$ & $234$ & $63\textrm{ns}$ & $936$B      \\
        Neural Network                  & $84.9\%$ & $3810$ & $635\textrm{ns}$ & $15$KB           \\
        Diff Logic Net (\textit{ours})   & $84.8\%$ & $1280$ & $5.1\textrm{ns}$ & $640$B           \\
    \toprule
        \textbf{Breast Cancer} & Acc. & \# Param. & Infer.~Time & Space \\
    \midrule
        Decision Tree Learner           & $71.9\%$ & \kern-2.3ex$\approx 100$ & $82$\textrm{ns} & \kern-2.3ex$\approx 230$B  \\
        Logistic Regression             & $72.9\%$ & $104$ & $34$ns & $416$B \\
        Neural Network                  & $75.3\%$ & $434$ & $130$ns & $1.4$KB   \\
        Diff Logic Net (\textit{ours})   & $76.1\%$ & $640$ & $2.8$ns & $320$B   \\
    \bottomrule
    \end{tabular}
    }
    \label{tab:adult-breast}
\end{wraptable}

On all three data sets, our method performs better than logistic regression and on MONK-3 (which is the data set with label noise) our method even outperforms the much larger neural network.
For hyperparameter details, see Supplementary Material~\ref{sm:model-architectures-hyperparameters}.

\subsection{Adult and Breast Cancer}

For our second set of experiments, we consider the Adult Census \cite{kohavi1996uci} and the Breast Cancer data set \cite{zwitter1988uci}.
We find that our method performs very similar to neural networks and logistic regression on the Adult data set while achieving a much faster inference speed.
On the Breast Cancer data set, our method achieves the best performance while still being the fastest model.
We present the results in Table~\ref{tab:adult-breast}.

\subsection{MNIST}

For our comparison to the fastest neural networks, we start by considering MNIST~\cite{lecun2010mnist}. 
The methods we compare ourselves to are also discussed in further detail in the baselines section.
In comparison to the fastest method achieving at least $98.4\%$ on MNIST, which is FINN by Umuroglu~\etal~\cite{umuroglu2017finn} (identified by Qin~\etal~\cite{qin2020binary}), our logic network achieves a better performance, while requiring less than $10\%$ of the number of binary operations.
That is, our model is objectively more than $10\times$ cheaper to evaluate.
When comparing real times, for an NVIDIA A6000 GPU, our model is $12\times$ faster than the model by Umuroglu~\etal~\cite{umuroglu2017finn} on their specialized FPGA hardware, even though our model only achieves a $7\%$ utilization of the GPU.
For the other BNNs, OPs have not been reported, but their inference speed is also substantially slower than FINN.
When compared to the smallest sparse neural network, our model requires substantially fewer operations than each of the of baselines.
Sparse function networks~\cite{gaier2019weight}, which have been learned evolutionarily, achieve an accuracy of $94.2\%$.
We provide an additional discussion of the results displayed in the Table~\ref{tab:mnist} in Supplementary Material~\ref{sec:difflogic:baselines}.

\begin{table}[h]
    \centering
    \vspace{-.75em}
    \caption{
        Results for MNIST, all of our results are averaged over $10$ runs. Times (T.) are inference times per image, the GPU is an NVIDIA A6000, and the CPU is a single thread at $2.5$~GHz. For our experiments, i.e., the top block, we use binarized MNIST.
    }
    \vspace{.5em}
    \scalebox{.94}{\footnotesize
    \begin{tabular}{llrrccrccc}
    \toprule
        \textbf{MNIST} & Acc. & \# Param. & Space & T.~[CPU] & T.~[GPU] & \textbf{OPs} & FLOPs  \\
    \midrule
        Linear Regression               & $91.6\%$ & $4\,010$           & $16$KB    & $3\mu$s       & $2.4$ns       & ($4$M)    & $4$K               \\  %
        Neural Network (\textit{small}) & $97.92\%$ & $118\,282$        & $462$KB   & $14\mu$s      & $12.4$ns      & ($236$M)  & $236$K                    \\
        Neural Network                  & $98.40\%$ & $22\,609\,930$    & $86$MB    & $2.2$ms       & $819$ns       & ($45$G)   & $45$M             \\
        \midrule
        Diff Logic Net (\textit{small})  & $97.69\%$ & $48\,000$         & $23$KB    & $625$ns       & $6.3$ns       & $48$K     & |                   \\
        Diff Logic Net                   & $98.47\%$ & $384\,000$        & $188$KB   & $7\mu$s       & ($50$ns)      & $384$K    & |                 \\
        \midrule
        \midrule
        \textit{Binary Neural Networks} &&&&& T.~[FPGA] \\
        FINN \cite{umuroglu2017finn}         & $98.40\%$ &              &           & ($96\mu$s)    & $641$ns       & $5.28$M   &                \\
        BinaryEye~\cite{jokic2018binaryeye}  & $98.40\%$ &              &           &               & $50\mu$s     &           &                  \\
        ReBNet~\cite{ghasemzadeh2018rebnet}  & $98.29\%$ &              &           &               & $3\mu$s     &           &                  \\
        LowBitNN~\cite{zhan2020field}        & $99.2\%$  &              &           &               & $152\mu$s     &           &                  \\
        \midrule
        \textit{Sparse Neural Networks} &&&&& Sparsity \\
        Var.~Dropout \cite{molchanov2017variational}    & $98.08\%$ & $4\,000$           &           &               & $98.5\%$      &  ($8$M)  & $8$K  &       \\
        $L_0$ regularization \cite{louizos2018learning} & $98.6\%$  &                    &           &               & $2/3$         &  ($200$M)  & $200$K          &        \\
        SET-MLP \cite{mocanu2018scalable}               & $98.74\%$ &  $89\,797$         &           &               & $96.8\%$      &  ($180$M)  & $180$K          &        \\
        \midrule
        Sparse Function Net \cite{gaier2019weight} & $94.2\%$ & $3\times1\,849$ &&&&& $>2$K \\  %
    \bottomrule
    \end{tabular}
    }
    \label{tab:mnist}
\end{table}

\subsection{CIFAR-10}

In addition to MNIST, we also benchmark our method on CIFAR-10~\cite{krizhevsky2009cifar10}.
For CIFAR-10, we reduce the color-channel resolution of the CIFAR-10 images and employ a binary embedding:
For a color-channel resolution of $4$ (the first three rows of Table~\ref{tab:cifar}), we use three binary values with the three thresholds $0.25, 0.5,$ and $0.75$.
For a color-channel resolution of $32$ (the large models, i.e., the last three rows of the top block of Table~\ref{tab:cifar}), we use 31 binary values with thresholds $(i/32)_{i\in\{1..31\}}$.
We do not apply data augmentation / dropout for our experiments, which could additionally improve performance.
For all baselines, we copied the reported accuracies from the original source, and thus those results are with data augmentation and the original color-channel resolution, and may include dropout \cite{srivastava2014dropout} as well as other techniques such as student-teacher learning with a convolutional teacher~\cite{urban2016deep}.

The results are displayed in Table~\ref{tab:cifar}.
We find that our method outperforms neural networks in the first setting (color-channel resolution of $4$) by a small margin, while requiring less than $0.1\%$ of the memory footprint and (with a larger model) by a large margin while requiring less than $1\%$ of the memory footprint.
In comparison to the best fully-connected neural network baselines, which are trained with various tricks such as student-teacher learning and retain the full color-channel resolution, our model does not achieve the same performance, while it is also much smaller and has access to fewer data.
With $1$ million parameters, the student-teacher model \cite{urban2016deep} has a footprint that is about $64\%$ larger than the footprint of our largest model (\textit{large}$\times4$), and achieves an accuracy which is only $3.7\%$ better than ours.
It is important to note that this models requires $2$ million floating-point operations, while our model requires $5$ million bit-wise logic operations (before pruning/optimization).
On float-arithmetic hardware-accelerated integrated circuits (as current GPUs and many CPUs), the $2$ million floating-point operations are around $100\times$ slower than $5$ million bit-wise logic operations.
On general purpose hardware (i.e., without float acceleration) the speed difference would be one order of magnitude larger, i.e., $1\,000\times$.

More competitive with respect to speed are sparse neural networks. 
In the final block of Table~\ref{tab:cifar}, we report the sparsest models for CIFAR-10.
Note that two of the methods resulted in performances below $50\%$, and even those methods which achieve around $75\%$ accuracy require a significantly more expensive inference.
ProbMask \cite{zhou2021effective} with its $140$ KFLOPs per image means that the model is (depending on hardware) $1-2$ orders of magnitude more expensive than our largest model.
Also, note that two out of three ResNet32 based sparsification methods achieve only around $37\%$ accuracy.

The results in parentheses are estimated because compilation to binaries did not finish / for GPU the largest models could also not be compiled due to compiler limitations. 
This can be resolved if desired with moderate implementation effort, e.g., compiling the model directly to PTX (CUDA Assembly) without compiling via \texttt{gcc} and \texttt{nvcc}.
The actual problem is that the compilers used by the implementation have a compile time that is quadratic in the number of lines of code / statements.

We provide an additional discussion of the results displayed in Table~\ref{tab:cifar} in SM~\ref{sec:difflogic:baselines}.

\begin{table*}[h]
    \centering
    \caption{
    Results on CIFAR-10. 
    Times (T.) are inference times per image, the GPU is an NVIDIA A6000, and the CPU is a single thread at $2.5$~GHz. For our experiments, i.e., the top block, we use a color-channel resolution of $4$ for the first $3$ lines and a color-channel resolution of $32$ for the \textit{large} models.
    The other baselines were provided with the full resolution of $256$ color-channel values.
    The numbers in parentheses are extrapolated / estimated.
    }
    \scalebox{.915}{\footnotesize
    \begin{tabular}{llrrccrccc}
    \toprule
        \textbf{CIFAR-10} & Acc. & \#\,Param & Space & T.~[CPU] & T.~[GPU] & \textbf{OPs} & FLOPs \\
    \midrule
        Neural Network (color-ch.~res.~= 4)        & $50.79\%$ & $12.6$M      & $48$MB    & $1.2$ms     & $370$ns  & ($25$G)    & $25$M     \\ %
        \midrule
        Diff Logic Net (\textit{small})              & $51.27\%$ & $48$K        & $24$KB    & $1.3\mu$s   & $19$ns   & $48$K      & |                          \\
        Diff Logic Net (\textit{medium})             & $57.39\%$ & $512$K       & $250$KB   & $7.3\mu$s   & $29$ns   & $512$K     & |                        \\
        Diff Logic Net (\textit{large})              & $60.78\%$ & $1.28$M      & $625$KB   & ($18\mu$s)  & ($73$ns) & $1.28$M    & |                         \\ %
        Diff Logic Net (\textit{large}$\times2$)     & $61.41\%$ & $2.56$M      & $1.22$MB  & ($37\mu$s)  & ($145$ns) & $2.56$M    & |                       \\
        Diff Logic Net (\textit{large}$\times4$)     & $62.14\%$ & $5.12$M      & $2.44$MB  & ($73\mu$s)  & ($290$ns) & $5.12$M    & |                       \\
        \midrule
        \midrule
        \textit{Best Fully-Connected Baselines~~(color-ch.~res.~= 256)}\kern-12em & \\
        Regularized SReLU NN \cite{mocanu2018scalable} & $68.70\%$  & $20.3$M        & $77$MB     & $1.9$ms     & $565$ns & ($40$G)    & $40$M           \\
        Student-Teacher NN \cite{urban2016deep}     & $65.8\%$  & $1$M                  & $4$MB      & $112\mu$s   & $243$ns & ($2$G)    & $2$M           \\
        Student-Teacher NN \cite{urban2016deep}     & $74.3\%$  & $31.6$M               & $121$MB    & $2.9$ms     & $960$ns & ($63$G)    & $63$M           \\
        \midrule
        \textit{Sparse Neural Networks} &&&&& Sparsity \\
        PBW (ResNet32) \cite{han2016deep}                           & $38.64\%$  &                &           &                     & $99.9\%$   & ($140$M)   & ($140$K)        \\ %
        MLPrune (ResNet32) \cite{zeng2019mlprune}                       & $36.09\%$  &                &           &                     & $99.9\%$   & ($140$M)   & ($140$K)       \\
        ProbMask (ResNet32) \cite{zhou2021effective}                      & $76.87\%$  &                &           &                     & $99.9\%$   & ($140$M)   & ($140$K)       \\
        SET-MLP \cite{mocanu2018scalable}                            & $74.84\%$  & $279$K         & $4.7$MB   &                     & $98.6\%$   & ($558$M)   & $558$K       \\
    \bottomrule
    \end{tabular}
    }
    \label{tab:cifar}
\end{table*}

\subsection{Distribution of Logic Gates}

\begin{figure*}[t]
    \centering
    \includegraphics[width=\linewidth, trim={0 9.5cm 0 .25cm},clip]{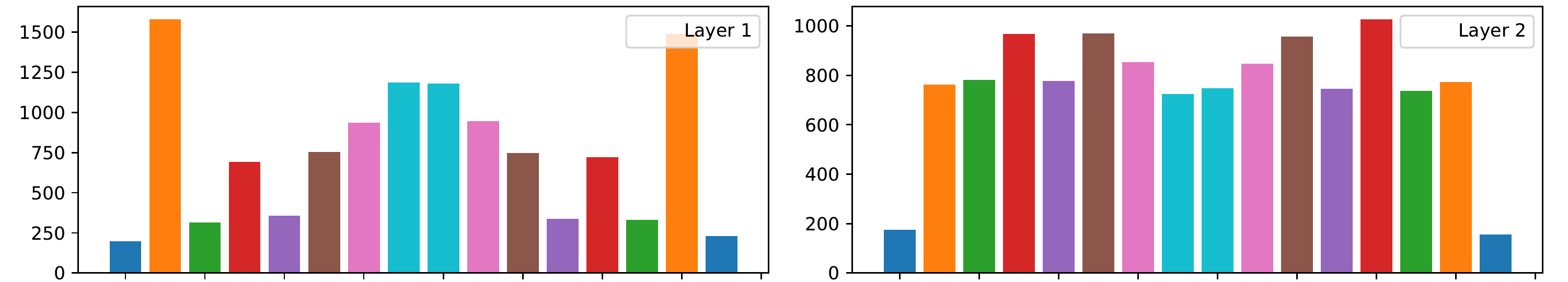}
    \includegraphics[width=\linewidth, trim={0 0 0 6.5cm},clip]{imgs/operator_distributions_00320550.pdf}
    \vspace{-1.5em}
    \caption{Distribution of logic gates in a trained four layer logic network.}
    \label{fig:operator-distributions}
\end{figure*}

To gain additional insight into learned logic gate networks, we consider histograms of operators present in each layer of a trained model.
Specifically, we consider a $4$ layer CIFAR-10 model with $12\,000$ neurons per layer in Figure~\ref{fig:operator-distributions}.

We observe that, generally, the constant $0/1$ ``operator'' is learned to be used only very infrequently as it does not actually provide value to the model.
Especially interesting is that it does not occur at all in the last layer.
In the first layer, we observe a stronger presence of `and', `nand', `or', and `nor'.
In the second and third layers, there are more `$A$', `$B$', `$\neg A$', and `$\neg B$'s, which can be seen as a residual / direct connection. 
This enables the network to model lower-order dependencies more efficiently by expressing it with fewer layers than the predefined number of layers.
In the last layer, the most frequent operations are `xor' and `xnor', which can create conditional dependencies of activations of the previous layers.
Interestingly, however, implications (e.g., $A\Rightarrow B$) are only infrequently used.

\section{Conclusion}

In this work, we presented a novel approach to train logic gate networks, which allows us to effectively train extremely efficient neural networks that---for their level of accuracy---are one or more orders of magnitude more efficient than the state-of-the-art.
For this, we leveraged real-valued logics and continuous relaxations via softmax.
We will release the source code of this work to the community to foster future research on learning logic gate networks. %

\begin{ack}
This work was supported by 
the IBM-MIT Watson AI Lab,
the DFG in the Cluster of Excellence EXC 2117 ``Centre for the Advanced Study of Collective Behaviour'' (Project-ID 390829875),
and the Land Salzburg within the WISS 2025 project IDA-Lab (20102-F1901166-KZP and 20204-WISS/225/197-2019).
\end{ack}

\printbibliography

\clearpage

\appendix

\section{Implementation Details}

We will release the source code of this work in form of a library building on top of PyTorch \cite{2019-PyTorch} and include the presented experiments for reproducibility.
For training, we use Python / PyTorch / CUDA, while for inference (apart from PyTorch) we use C for CPU and CUDA for GPU.
The library supports both a pure PyTorch implementation as well as a native CUDA implementation (which is up to $50\times$) faster than the pure PyTorch implementation. 
Both implementations are usable in PyTorch: pure PyTorch / native CUDA just refers to the backend.
For inference, our library automatically converts trained logic gate network into optimized C / CUDA binaries, allowing for easy and fast deployment, callable directly from Python (but also from any other language that can handle shared object binaries.)

\subsection{Model Architectures and Hyperparameters}
\label{sm:model-architectures-hyperparameters}

For all experiments, we used ``straight'' architectures, i.e., architectures with the same number of neurons per layer.
In Tables~\ref{tab:network-architectures-logic} and~\ref{tab:network-architectures-mlp}, we display the numbers of layers numbers of neurons per layer for each network architecture used in this work.
In general, the architecture search for all models was performed via grid search with the number of layers in $\{2, 3, 4, 5, 6, 7, 8, 9, 10\}$ and number of neurons per layer with a resolution of factor $2$.

For all models, we use the Adam optimizer~\cite{kingma2015adam}.
For all neural networks, we use a learning rate of $0.001$ and for all logic gate networks, we use a learning rate of $0.01$.
We train all models up to $200$ epochs at a batch size of $100$.
The softmax temperature $\tau$ was searched over a grid of $\{1, 1/0.3, 1/0.1, 1/0.03, 1/0.01\}$ (except for Adult). 
As the experiments for were originally parameterized via an inverse temperature $1/\tau$, we provide the exact fractions to prevent rounding errors.
The optimal temperature primarily depends on the number of outputs per class.
If there are more outputs per class, the range of predictions is larger, and thus, we use a larger temperature to counter this effect.

\begin{table*}[h]
    \centering
    \caption{Logic gate network architectures.}
    {\footnotesize
    \begin{tabular}{lcrrrr}
    \toprule
        Dataset     & Model     & Layers & Neurons / layer & Total num.~of p. & $\tau$  \\
    \midrule
        MONK-1      & ---       & $6$ & $24$ & $144$ & $1$ \\
        MONK-2      & ---       & $6$ & $12$ & $ 72$ & $1$ \\
        MONK-3      & ---       & $6$ & $12$ & $ 72$ & $1$ \\ 
    \midrule
        Adult       & ---       & $5$ & $256$ & $1\,280$ & $1/0.075$\\
        Breast Cancer & ---     & $5$ & $128$ & $640$ & $1/0.1$ \\
    \midrule
        MNIST       & small     & $6$ & $8\,000$ & $48\,000$ & $1/0.1$\\
                    & normal    & $6$ & $64\,000$ & $384\,000$ & $1/0.03$ \\
    \midrule
        CIFAR-10    & small     & $4$ & $12\,000$  & $48\,000$ & $1/0.03$ \\
                    & medium    & $4$ & $128\,000$ & $512\,000$ & $1/0.01$ \\
                    & large     & $5$ & $256\,000$ & $1\,280\,000$ & $1/0.01$ \\
                    & large$\times2$ & $5$ & $512\,000$ & $2\,560\,000$ & $1/0.01$ \\
                    & large$\times4$ & $5$ & $1\,024\,000$ & $5\,120\,000$ & $1/0.01$ \\
    \bottomrule
    \end{tabular}
    }
    \label{tab:network-architectures-logic}
\end{table*}

\begin{table*}[h]
    \centering
    \caption{Multi-layer perceptron / neural network baseline architectures. All architectures are ReLU activated.}
    {\footnotesize
    \begin{tabular}{lcrrr}
    \toprule
        Dataset     & Model     & Layers & Neurons / layer & Total num.~of parameters  \\
    \midrule
        MONK-1      & ---       & $2$ & $8$ & $162$ \\
        MONK-2      & ---       & $2$ & $8$ & $162$ \\
        MONK-3      & ---       & $2$ & $8$ & $162$ \\ 
    \midrule
        Adult       & ---       & $2$ & $32$ & $3\,810$ \\
        Breast Cancer & ---     & $2$ & $8$ & $434$ \\
    \midrule
        MNIST       & small     & $3$ & $128$ & $118\,282$ \\
                    & normal    & $7$ & $2\,048$ & $22\,609\,930$ \\
    \midrule
        CIFAR-10    & ---       & $5$ & $1\,024$ & $12\,597\,258$ \\
    \bottomrule
    \end{tabular}
    }
    \label{tab:network-architectures-mlp}
\end{table*}

\subsection{Aggregating Predictions via Logic Gate Network Adders}
Optionally, we aggregate the output bits for each class into a binary number to reduce the required memory bandwidth for returning the predictions.
This is done after learning and can be expressed via a fixed logic gate network. 
Specifically, we implement adders, which can add one bit to a binary number with logic gates. 
This way, the aggregation is extremely efficient, specifically, the aggregation is faster than storing the un-aggregated results in the VRAM.

\subsection{Training Times \& Standard Deviations}

Here, we provide training times for the MNIST and CIFAR-10 models.
The training times are for the version of the code used for the original experiments. We will also make a substantially faster implementation publicly available.
In addition, we provide the standard deviations for the accuracy.

\begin{table*}[h]
    \centering
    \caption{Training times and standard deviations for the experiments on MNIST.}
    {\footnotesize
    \begin{tabular}{lcc}
    \toprule
        Model & Training Time & Accuracy \\
    \midrule
        Neural Net Baseline (\textit{small})   & 0.3 h & $97.92\% \pm 0.08\%$ \\
        Neural Net Baseline           & 0.4 h & $98.40\% \pm 0.06\%$ \\
    \midrule
        Diff Logic Net (\textit{small})        & 1.8 h & $97.69\% \pm 0.11\%$ \\
        Diff Logic Net                & 5.3 h & $98.47\% \pm 0.05\%$ \\
    \bottomrule
    \end{tabular}
    }
    \label{tab:training-time-std-mnist}
\end{table*}

\begin{table*}[h]
    \centering
    \caption{Training times and standard deviations for the experiments on CIFAR-10.}
    {\footnotesize
    \begin{tabular}{lcc}
    \toprule
        Model & Training Time & Accuracy \\
    \midrule
        Neural Net Baseline       & 0.8 h  & $50.79 \pm 0.35$ \\
    \midrule
        Diff Logic Net (\textit{small})    & 1.3 h  & $51.27 \pm 0.26$ \\ 
        Diff Logic Net (\textit{medium})   & 7.4 h  & $57.39 \pm 0.13$ \\
        Diff Logic Net (\textit{large})    & 24.2 h & $60.78 \pm 0.12$ \\
        Diff Logic Net (\textit{large}$\times2$)         & 45.6 h & $61.41 \pm 0.02$ \\ 
        Diff Logic Net (\textit{large}$\times4$)         & 90.3 h & $62.14 \pm 0.02$ \\
    \bottomrule
    \end{tabular}
    }
    \label{tab:training-time-std-cifar-10}
\end{table*}

\section{Additional Discussion of Baselines}
\label{sec:difflogic:baselines}

In this section, we provide an additional discussion of fast network architectures as baselines for differentiable logic gate nets.

\subsection{Binary Neural Networks}

Qin~\etal~\cite{qin2020binary} give a current overview of binary neural networks in their survey. 
They discuss the challenges of training binary neural networks or translating existing neural networks into their binarized counterparts.
They identify FINN by Umuroglu~\etal~\cite{umuroglu2017finn} as the fastest method for classifying MNIST at an accuracy of $98.4\%$ at a frame rate of $1\,561\,000$ images per second on specialized FPGA hardware.

FPGAs (field-programmable gate arrays) are configurable hardware accelerated processors that can achieve extreme speeds for fixed, predefined tasks that are expressed via logic gates.
As FPGAs operate at extreme speeds, they were also used for applications such as mining cryptocurrencies~\cite{tetu2020standalone, Kolivas2013cgminer}
or even implementing an oscilloscope \cite{khan2005implementation}
since, here, the required complexity is rather limited, while high speeds are necessary.

The binary FINN MNIST model by Umuroglu~\etal~\cite{umuroglu2017finn} requires $5.82$ MOPs (Mega binary OPerations) per frame, which means that their FPGA achieves around $5.82\cdot10^6\cdot1.561\cdot10^6 = 9.09 \cdot10^{12}$ binary operations per second, i.e., $9.09$ TOPS (Tera binary OPerations per Second).
Conventional CPUs, are $10-100$ times slower than their FPGA.
On different FPGAs, Ghasemzadeh~\etal~\cite{ghasemzadeh2018rebnet} achieve $330\,000$ images per second on MNIST at an accuracy of $98.29\%$, and Jokic~\etal~\cite{jokic2018binaryeye} propose an FPGA based embedded camera system achieving $20\,000$ images per second at $98.4\%$ accuracy.

Zhan~\etal~\cite{zhan2020field} concentrate on deploying Low-Bit Neural Networks (LBNNs) on FPGAs and achieve an accuracy of $99.2\%$ on MNIST at $6\,580$ images per second. 
Shani~\etal~\cite{shani2019dynamics} explore analog logic gate nets (a physical approximation to Boolean nets) and achieve accuracies up to $89\%$ on MNIST.

\subsection{Sparse Neural Networks}

Sparse neural networks are neural networks where only a selected subset of connections is present, i.e., instead of fully-connected layers, the layers are \textit{sparse}.
Hoefler~\etal~\cite{hoefler2021sparsity} give an overview of sparsity on deep learning in their recent literature review.
They identify Molchanov~\etal~\cite{molchanov2017variational} to achieve the sparsest (originally fully-connected) model on MNIST with a sparsity of $98.5\%$ achieving an accuracy of $98.08\%$.
For this, Molchanov~\etal~\cite{molchanov2017variational} propose variational dropout with unbounded dropout rates to sparsify neural networks.
Louizos~\etal~\cite{louizos2018learning} propose sparsification via $L_0$ regularization and report an MNIST accuracy of $98.6\%$ for a model with around $2\cdot 10^{5}$~FLOPs (FLoating point OPerations), which corresponds to a sparsity of around $2/3$.

Zhou~\etal~\cite{zhou2021effective} propose ProbMask and give an overview over the sparsest CIFAR-10 models.
They report up to a sparsity of $99.9\%$, which corresponds to around $140$ kFLOPs for their smallest network architecture (ResNet32, which has a base cost of $~140$~MFLOPs \cite{lym2019prunetrain}.) %
This is the only work where architectures this sparse are reported in the literature.
While these models have the advantage of being based on the VGG and ResNet CNN architectures, our models are still very competitive, especially considering that our models are much smaller than their smallest reported results.

Blalock~\etal~\cite{blalock2020state} report in their survey that for CIFAR-10 with a VGG, to achieve a theoretical speedup of $32\times$, all evaluated methods drop significantly below $70\%$ test accuracy.
Note that this speedup corresponds to a sparsity of around $97\%$, which makes up a much larger model than the models considered in this work.

Mocanu~\etal~\cite{mocanu2018scalable} propose training neural networks with sparse evolutionary training inspired by network science.
Their method evolves an initial sparse topology of two consecutive layers of neurons into a scale-free topology.
They achieve an accuracy of $74.84\%$ on CIFAR-10 with $278\,630$ floating-point parameters.
We estimate this to correspond to a theoretical cost of around $550$ kFLOPs (multiplication + addition), corresponding to $550$ MOPs as per our conservative estimate.
On MNIST, they achieve (with $89\,797$ parameters) an accuracy of $98.74\%$.
Thus, the model is by orders of magnitude more expensive to evaluate than the logic gate networks considered in this work.

A FLOP generally corresponds to many binary OPs.
Specifically, a float32 adders / multiplier requires usually at least $1\,000$ logical gates or look up tables and usually has a delay of tens of logical levels.
Practically, float32 adders / multipliers are implemented directly in hardware in CPUs and GPUs, as it is an essential operation on such platforms.
Nevertheless, also in practice, a float32 adder / multiplier is much more expensive than performing a bitwise logical operation on int64 data types (even on float32 and int32 focussed GPUs). 
On CPUs, around $3-10$ int64 bit-wise operations can be performed per cycle, while floating-point operations usually require a full clock cycle.
To convert a non-sparse model we assume a very conservative $100$ OPs per $1$ FLOP.
Note that speeds for sparse neural networks are also only theoretical because sparse execution usually brings an overhead of factor $10-100\times$.
So overall, in practice, $1\,000$ (binary) OPs per $1$ sparse (float32) FLOP is a very conservative estimate in favor of sparse float32 models.
Further, in theory, $1\,000$ OPs per $1$ FLOP is an accurate estimate (assuming sparsity to come without cost and assuming floating-point operations to not be hardware accelerated).

\section{Additional Discussions}
\subsection{Depth vs.~Accuracy Trade-off}
We observed a trade-off between depth and accuracy that is similar to the trade-off for regular neural networks.
In our experiments, we found that logic gate networks can generally be trained efficiently up to around 8-10 layers, when training starts to suffer from vanishing gradients. 
This is similar to where vanishing gradients start to be a problem in regular neural networks, at least without applying tricks like residual connections or batch norm. 

\subsection{On the Effectiveness of Randomized Sparse Connections}
\vspace{-.25em}

We rely on fixed connections because learning the connectivity would require a relaxed connectivity, which would add additional complexity to the relaxation, which would degrade performance. 
However, we note that updating the connectivity based on some heuristic after a certain amount of training could, in principle, improve performance. 
This could be a subject of future work.

Sparsity can actually be viewed from two related angles: 
first, sparsity arises from the definition of binary logic gate operators leading to each neuron having only two inputs, which contrasts regular fully connected networks, where each neurons is a weighted sum of all inputs; 
second, sparsity can be seen from the perspective of the number of pairs of neurons covered: 
here, for $n$ inputs to a layer, we have $n\cdot (n+1)/2$ possible pairs of inputs, but typically only choose to consider $n$ pairs to avoid an exploding number of neurons in the downstream layers. 
Here, we use a random selection of connections as it is the canonical choice.
We found that, as long as not only neighboring pairs of neurons are selected, the method of selection does not substantially affect performance, which is why we stuck with random connections to simplify the method.
In future work, speed improvements (training and inference) could be possible by designing sparsity patterns that lead to faster memory access on respective hardware, while keeping the selection sufficiently ``random'' or ``shuffled'' such that accuracy is not impacted.
As to why sparse and random connections work well in the first place, Liu~\etal~\cite{liu2022unreasonable} discuss and investigate randomly selected sparse connections in regular neural network in great detail and demonstrate their effectiveness.

\vspace{-.25em}
\subsection{Ternary and Other Additional Operators}
\vspace{-.25em}
In this work, we focus on binary logic gates. 
However, one may also consider ternary logic gates, i.e., logic gates with three inputs, e.g., $a\wedge b \wedge c$, or the more general form of $k$-ary logic gates.
For $k$ binary inputs there are exactly $2^{(2^k)}$ possible binary operators. 
Thus, e.g., for $3$ inputs, there are $256$ possible binary operators. 
An important reason to limit the number of possible operators is that too many operators would lead to vanishing probabilities, thereby inhibiting training. 
Further, additional operators would lead to computationally more expensive training because more relaxed logic operators would need to be computed, more outputs would need to be aggregated, and more derivatives would need to be computed.

\vspace{-.5em}
\section{Differentiable Logics: T-Norms and T-Conorms}
\vspace{-.5em}
\label{sm:tcn}

Here, we cover various T-norms and T-conorms, which are the build blocks of real-valued logics, and could be considered as alternatives to the probabilistic T-norm and T-conorm used in the main paper.

The axiomatic approach to multi-valued logics (which we need to combine
the occlusions by different faces in a ``soft'' manner) is based on
defining reasonable properties for truth functions. We state the axioms
for multi-valued generalizations of the conjunction (logical
``and''), called T-norms, in Definition~\ref{def:t-norm} and generalizations of the disjunction (logical
``or''), called T-conorms, in Definition~\ref{def:t-conorm}. 
\begin{definition}[T-norm]\label{def:t-norm}
A T-norm (triangular norm) is a binary
operation~$\top: [0,1] \times [0,1] \to [0,1]$,
which satisfies
\begin{itemize}\itemsep0pt
\item associativity:
      $\top(a, \top(b,c)) = \top(\top(a,b), c)$,
\item commutativity:
      $\top(a,b) = \top(b,a)$,
\item monotonicity:
      $(a \leq c) \land (b \leq d)
       \Rightarrow \top(a,b) \leq \top(c,d)$,
\item $1$ is a neutral element:
      $\top(a,1) = a$.
\end{itemize}
\end{definition}

\begin{definition}[T-conorm]\label{def:t-conorm}
    A T-conorm is a binary operation~$\bot : [0,1] \times [0,1] \to [0,1]$, which satisfies
    \begin{itemize}
    \setlength\itemsep{0em}
        \item associativity: $\bot(a, \bot(b, c)) = \bot(\bot(a, b), c)$,
        \item commutativity: $\bot(a, b) = \bot(b, a)$,
        \item monotonicity: $(a\leq c) \land (b\leq d) \Rightarrow \bot(a, b) \leq \bot(c, d)$,
        \item $0$ is a neutral element $\bot(a, 0) = a$.
    \end{itemize}
\end{definition}
\begin{remark}[T-conorms and T-norms]
    While T-conorms~$\bot$ are the real-valued equivalents of the logical `or',
    so-called T-norms~$\top$ are the real-valued equivalents of the logical `and'. %
    Certain T-conorms and T-norms are dual in the sense that
    one can derive one from the other using a complement (typically $1-x$) and De~Morgan's laws ($\top(a, b) = 1-\bot(1-a, 1-b)$).
\end{remark}

Clearly, these axioms ensure that the corners of the unit square, that
is, the value pairs considered in classical logic, are processed as
with a standard conjunction: neutral element and commutativity imply
that $(1,1) \mapsto 1$, $(0,1) \mapsto 0$, $(1,0) \mapsto 0$. From
one of the latter two and monotonicity it follows $(0,0) \mapsto 0$.
Analogously, the axioms of T-conorms ensure that the corners of the
unit square are processed as with a standard disjunction. Actually,
the axioms already fix the values not only at the corners, but on
the boundaries of the unit square. Only inside the unit square (that
is, for $(0,1)^2$) T-norms (as well as T-conorms) can differ.

In the theory of multi-valued logics, and especially in fuzzy logic \cite{klir1997fuzzy},
it was established that the largest possible T-norm is the minimum
and the smallest possible T-conorm is the maximum: for any
T-norm~$\top$ it is $\top(a,b) \le \min(a,b)$ and for any
T-conorm~$\bot$ it is $\bot(a,b) \ge \max(a,b)$. The other extremes,
that is, the smallest possible T-norm and the largest possible T-conorm
are the so-called drastic T-norm, defined as $\top^\circ(a,b) = 0$
for $(a,b) \in (0,1)^2$, and
the drastic T-conorm, defined as
$\bot^\circ(a,b) = 1$ for $(a,b) \in (0,1)^2$. Hence, it is
$\top(a,b) \ge \top^\circ(a,b)$ for any T-norm~$\top$ and
$\bot(a,b) \le \bot^\circ(a,b)$ for any T-conorm~$\bot$.
We do not consider the drastic T-conorm for an occlusion test because it clearly does not
yield useful gradients.

As mentioned, it is common to combine a
T-norm~$\top$, a T-conorm~$\bot$ and a negation~$N$ (or complement,
most commonly $N(a) = 1-a$) so that DeMorgan's laws hold. Such a
triplet is often called a {\em dual triplet}.
In Tables~\ref{tab.t-norms} and~\ref{tab.t-conorms} we show the
formulas for the families of T-norms and T-conorms, respectively,
which, together with the standard negation $N(a) = 1-a$,
form dual triplets. 

Finally, we would like to recapitulate that, in this work, we used the probabilistic T-norm / T-conorm.

\begin{table}[h]
\centering
\caption{\label{tab.t-norms}(Families of) T-norms.}
\vspace{.5em}
\begin{tabular}{|l|c@{~$=$~}l|}\hline\rule{0pt}{2.6ex}%
Minimum         & $\top^M(a,b)$
  & $\min(a,b)$                  \\[1.2ex]
Probabilistic   & $\top^P(a,b)$
  & $ab$                         \\[1.2ex]
Einstein        & $\top^E(a,b)$
  & $\frac{ab}{2-a-b+ab}$        \\[1.2ex]
Hamacher        & $\top^H_p(a,b)$
  & $\frac{ab}{p+(1-p)(a+b-ab)}$ \\[1.2ex]
Frank           & $\top^F_p(a,b)$
  & $\log_p\left(1+\frac{(p^a-1)(p^b-1)}{p-1}\right)$ \\[1.2ex]
Yager           & $\top^Y_p(a,b)$
  & $\max\left(0, 1-\left(\left(1-a\right)^p+\left(1-b\right)^p\right)^{\frac{1}{p}}\right)$ \\[1.2ex]
Acz\'el-Alsina  & $\top^A_p(a,b)$
  & $\exp\big(-\left(|\log(a)|^p+|\log(b)|^p
               \right)^{\frac{1}{p}}\big)$ \\[0.3ex]
Dombi           & $\top^D_p(a,b)$
  & $\Big(1+\left( \left(\frac{1-a}{a}\right)^p
                  +\left(\frac{1-b}{b}\right)^p
           \right)^{\frac{1}{p}}\Big)^{\!\!-1}$ \\[1.2ex]
Schweizer-Sklar & $\top^S_p(a,b)$
  & $(a^p+b^p-1)^{\frac{1}{p}}$ \\[1ex] \hline
\end{tabular}
\end{table}

\begin{table}[h]
\centering
\vspace{-.5em}
\caption{\label{tab.t-conorms}(Families of) T-conorms.}
\vspace{.5em}
\begin{tabular}{|l|c@{~$=$~}l|}\hline\rule{0pt}{2.6ex}%
Maximum         & $\bot^M(a,b)$
  & $\max(a,b)$                  \\[1.2ex]
Probabilistic   & $\bot^P(a,b)$
  & $a+b-ab$                     \\[1.2ex]
Einstein        & $\bot^E(a,b)$
  & $\bot^H_2(a,b)=\frac{a+b}{1+ab}$           \\[1.2ex]
Hamacher        & $\bot^H_p(a,b)$
  & $\frac{a+b+(p-2)ab}{1+(p-1)ab}$ \\[1.2ex]
Frank           & $\bot^F_p(a,b)$
  & $1-\log_p\left(1+\frac{(p^{1-a}-1)(p^{1-b}-1)}{p-1}\right)$ \\[1.2ex]
Yager           & $\bot^Y_p(a,b)$
  & $\min\left(1, (a^p+b\kern0.1pt^p)^{\frac{1}{p}}\right)$ \\[1.2ex]
Acz\'el-Alsina  & $\bot^A_p(a,b)$
  & $1 -\exp\big(-\left(|\log(1-a)|^p+|\log(1-b)|^p
                 \right)^{\frac{1}{p}}\big)$ \\[0.3ex]
Dombi           & $\bot^D_p(a,b)$
  & $\Big(1+\left( \left(\frac{1-a}{a}\right)^p
                  +\left(\frac{1-b}{b}\right)^p
           \right)^{-\frac{1}{p}}\Big)^{\!\!-1}$ \\[1.2ex]
Schweizer-Sklar & $\bot^S_p(a,b)$
  & $1-((1-a)^p+(1-b)^p-1)^{\frac{1}{p}}$ \\[1ex] \hline
\end{tabular}
\vspace{-1em}
\end{table}

\end{document}